\documentclass{article}
\usepackage{spconf,amsmath,graphicx}
\usepackage{cite}
\usepackage{amsmath,amssymb,amsfonts}
\usepackage{amsmath}
\usepackage{algorithm}
\usepackage{algpseudocode}
\usepackage{graphicx}
\usepackage{textcomp}
\def\BibTeX{{\rm B\kern-.05em{\sc i\kern-.025em b}\kern-.08em
    T\kern-.1667em\lower.7ex\hbox{E}\kern-.125emX}}

\DeclareMathOperator*{\argmin}{argmin}

\title{CHANNEL PRUNING IN QUANTIZATION-AWARE TRAINING: AN ADAPTIVE PROJECTION-GRADIENT DESCENT-SHRINKAGE-SPLITTING METHOD}
\name{Zhijian Li$^{\star}$, \qquad Jack Xin$^{\star}$\thanks{This work is partly supported by NSF grants DMS-1854434, DMS-1952644, DMS-1924548.}}
\address{$^{\star}$University of California, Irvine\\
Department of Mathematics}
\begin{document}
%
\maketitle
\begin{abstract}
We propose an adaptive projection-gradient descent- shrinkage- splitting method (APGDSSM) to integrate penalty based channel pruning into quantization-aware training (QAT). APGDSSM concurrently searches  weights in both the quantized subspace and the sparse subspace. APGDSSM uses shrinkage operator and a splitting technique to create sparse weights, as well as the Group Lasso penalty to push the weight sparsity into channel sparsity. In addition, we propose a novel complementary transformed $l_1$ penalty to stabilize the training for extreme compression.
\end{abstract}
\section{Introduction}
 Convolutional neural networks (CNNs) have been widely used for computer vision tasks such as image classification and segmentation. To increase efficiency and reduce memory costs in mobile and IoT applications, network compression is necessary. Quantization and channel pruning are two commonly adopted methods. QAT searches the optimal weight in the quantized subspace.
For a CNN with $L$ convolutional layers, let $\mathbf{w}= \{w_1, \cdots, w_L\}$ be weight tensors structured in (height, width, channel) per layer. 
The subspace of $m$-bit ($m\geq 2$) quantization $\mathcal{Q}\in R^n$ is 
$$\mathcal{Q}=\mathbb{R}\times \{0,\pm 1,\pm 2, \cdots, \pm 2^{m-1}\}^n$$
Given an objective function $\mathcal{L}$, the quantization problem  is $\argmin_{\mathbf{u}\in \mathcal{Q}}\mathcal{L}(\mathbf{u})$
for which \cite{binaryconnect} proposed a widely used QAT algorithm based on an auxiliary float weight $w$ to perform QAT. With learning rate $\gamma$, it can be formulated as
\begin{equation}
    \mathbf{w} \leftarrow \mathbf{w}-\gamma \nabla \mathcal{L}(\mathbf{u}),\;
    \mathbf{u} \leftarrow \text{Proj}_{Q}\, (\mathbf{w})
    \label{eq:bc}
\end{equation}
where the $\text{Proj}_{Q}(\cdot)$ is the projection that maps the float weight into the quantized subspace.  For a  theoretical convergence analysis of \eqref{eq:bc} and a relaxed formulation with improved performance , see  \cite{binaryrelax}.
Channel pruning is a structured compression well-studied by itself ( \cite{wen2016learning,LiuDar_19,bui2020nonconvex,yang2019channel,dinh2020sparsity} and references therein). Integrating QAT into adversarial training and studying the  sparsity of quantized models are performed in \cite{li2021}
\medskip

The main contribution of our work here is to propose an integrated objective to {\it do channel pruning and weight quantization in one shot}. This is achieved by minimizing a new objective function with group sparse penalty 
over $Q$ through an adaptive splitting, projection, gradient descent and proximal operations (APGDSSM algorithm). The adaptive step is to avoid weights in a layer all becoming very small, or fix potential model collapse when trained by the integrated steps of the algorithm. Besides adapting training schedule, we also found a new penalty, the so called complementary 
transformed-$\ell_1$ (CT$\ell_1$), to steer weights away from the trivial state in each layer. Using CT$\ell_1$)
gives more room to trade-off 
accuracy for efficiency than adapting training schedule. 
Experimental results on CIFAR-10, CIFAR-100, and Imagenet support our proposed methodology and framework.

\section{Related Work}
For a loss function $l$, the Lasso regularized problem is  
\begin{equation}
\mathcal{L}(\mathbf{w})=l(\mathbf{w})+\lambda ||\mathbf{w}||_1.
\label{eq:lasso}
\end{equation}
It is well-known that Lasso regularization does parameter selection for the model, and 
several approaches exist for solving problem \eqref{eq:lasso}. 
In \cite{fista}, an iterative algorithm of proximal operator (FISTA) solves \eqref{eq:lasso}, where the proximal operator for a penalty function $g$ is defined as $\text{Prox}_{g}(\mathbf{w})=\argmin_{\mathbf{u}}g(\mathbf{u})+\frac{1}{2}||\mathbf{u}-\mathbf{x}||^2$. The algorithm is: 
$$\mathbf{w}^{t+1}= \text{Prox}_{\lambda}\big(\mathbf{w}^t-\gamma \nabla f(\mathbf{w}^t)\big)$$
where  
$$\text{Prox}_{\lambda}(x)={\rm sgn}(x)\cdot \max{(|x|-\lambda, 0)}.$$
An alternative method to solve \eqref{eq:lasso} is the  Alternating Direction Method of Multipliers (ADMM), through an augmented Lagrangian (Boyd et al. \cite{admm}):
\begin{equation}
    \mathcal{L}(\mathbf{w},\mathbf{u},\mathbf{z})=f(\mathbf{w})+\lambda ||\mathbf{u}||_1+\langle \mathbf{z}, \mathbf{w}-\mathbf{u} \rangle+\frac{\beta}{2}||\mathbf{w}-\mathbf{u}||^2
\end{equation}
ADMM is adapted 
to neural network training in  \cite{taylor2016training,
ye2019adversarial}.
The convergence theorems of ISTA and ADMM require both the loss function and penalty function to be convex, which does not apply to deep neural networks. The relaxed splitting variable method (RSVM,\cite{rvsm})  sparsifies non-convex neural networks by minimizing a simplified augmented Lagrangian:
$$l_{\lambda/\beta}(\mathbf{w},\mathbf{u})= f(\mathbf{w})+\lambda ||\mathbf{u}||_1+\frac{\beta}{2} ||\mathbf{u}-\mathbf{w}||^2.$$
RVSM updates weights as 
\begin{equation}
    \mathbf{w} \leftarrow \mathbf{w} - \gamma \nabla f(\mathbf{w}) - \gamma \beta(\mathbf{w}-\mathbf{u}), \;
    \mathbf{u} \leftarrow \text{Prox}_{\lambda/\beta}(\mathbf{w})
\end{equation}
which extends to non-differential penalties (e.g. $\ell_0$) with the corresponding proximal operator.
The RVSM does not require convex or differentiable penalty function for convergence \cite{rvsm}, and it applies  to  adversarially trained networks \cite{dinh2020sparsity}.
Though models trained by RVSM usually have unstructured sparsity with limited channel sparsity, 
 RVSM extends readily to a group-wise variable splitting method (RGSM, \cite{yang2019channel}) based on Group Lasso (GL) penalty:
$$||\mathbf{w}||_{GL}=\sum_{l=1}^{L}\sum_{i\in I_l}||w_{l,i}||_2$$
to increase channel sparsity, 
where $I_l$ is the collection of channels in the $l$-th layer.
GL penalty with its proximal operator in closed form is applied channel-wise in network training to 
realize 
sparse channels \cite{yang2019channel,dinh2020sparsity}. In \cite{LS_20}, RGSM and QAT are combined in a multi-stage process to achieve both channel pruning and 
binary weights.

\section{Methodology and APGSSM Algorithm}
To train quantized neural networks with sparse channels, we proposed an algorithm to concurrently search the optimal weights in the quantized subspace and the sparse subspace, as shown in Algorithm 1. The objective is
\begin{equation}
    \min_{u\in \mathcal{Q}}\;  \mathcal{L}(\mathbf{u}): =l(\mathbf{u})+\lambda_2||\mathbf{u}||_{GL}+\lambda_1||\mathbf{u}||_1
    \label{eq:obj1}
\end{equation}
The procedure of training is shown in Algorithm 1. We note that the Lasso regularization term in equation \eqref{eq:obj1} is imposed implicitly. As shown in Algorithm 1, the $l_1$ penalty does not contribute to the gradient. Instead, we use the shrinkage operator to minimize it. For parameters, we use symbols against the epoch number $t$, e.g. $\lambda_1^t$, to indicate that there is an adaptive scheme for the values.

\begin{algorithm}
\caption{APGDSM and APGDSSM}\label{alg:cap}
\hspace*{\algorithmicindent} \textbf{Input}: Float weights $w^0$. Hyperparameters $\lambda_1, \lambda_2, \beta$.  \\
\hspace*{\algorithmicindent} \textbf{Output} : Quantized weights $u$.
\begin{algorithmic}
\For{$t = 1,\cdots, 200$}:
    
    \State $\mathbf{u}^t = Proj_{Q}(\mathbf{w}^t)$
    \State $f(\mathbf{u}^t)=l(\mathbf{u}^t)+\lambda_2^t||\mathbf{u}^t||_{GL}$
    \State $\mathbf{w}^{t}=\mathbf{w}^{t-1}-\alpha\nabla f(\mathbf{u}^t)$ 
    \If{Splitting}: \Comment{Split if APGDSSM}
        \State $\mathbf{w}^t=\mathbf{w}^t-\gamma^t \beta^t (\mathbf{w}^t-\mathbf{u}^t)$
    \EndIf
    \State $\mathbf{w}^{t}=Prox_{ \lambda_1^t}(\mathbf{w}^{t}_g)$
\EndFor
\State $\mathbf{u} = Proj_{Q}(\mathbf{w}^{200})$
\end{algorithmic}
\label{alg:algorithm1}
\end{algorithm}
This algorithm concurrently searches both the quantized subspace and the subspace of sparse weight (with small $l_1$ norm). We can either use only shrinkage operator (APGDSM) or use it together with the splitting (APGDSSM). The splitting term 
updates the gradient descent of $\frac{\beta}{2}||\mathbf{w}^t-\mathbf{u}^t||^2$, which makes the float weight $\mathbf{w}^t$ close to the quantized weight $\mathbf{u}^t$. Since $\mathbf{u}^t$ is much more sparse than $\mathbf{w}^t$, the splitting step renders $\mathbf{w}^t$ with more small elements, which strengthens the performance of the following shrinkage operator. However, pushing $\mathbf{w}^t$ close to $\mathbf{u}^t$ can jeopardize the performance, as it is not the descending direction guided by gradient.
\section{Implementation and Experiments}
 \begin{table}[ht!]
     \centering
     \begin{tabular}{|c|c|c|}
     \hline
         Epoch &  Factor for $\lambda_1\&\lambda_2$& Factor for $\beta$\\
         \hline
         35 & 0.5&0.5\\
         70 & 0.2&0.2\\
         110&0.5&0.1\\
         150&0.5&0.1\\
         \hline
     \end{tabular}
     \caption{Adaptive scheme for the parameters in Algorithm 1. At epochs listed in the left-side column, we multiply the parameters by the fatcor in the right-side column}
     \label{tab:scheme}
 \end{table}
 
We use the standard adaptive scheme for the learning rate $\gamma^t$. The initial learning rate is $0.1$, and we multiply the learning rate by a factor of $0.1$ at epochs $80$, $120$, and $160$. During the training, we need to change the scale of the regularization parameters to fit the current learning rate. For both $\lambda_1$, $\lambda_2$, and $\beta$, we empirically design a scheme to adapt the values of parameters. The reason we have a different adaptive scheme from the learning rate is that the training has a high probability to collapse if the parameters are re-scaled too late. As in Algorithm 1, all GL regularization, shrinkage operator, and splitting terms drive the weights to be sparse. When this force of sparsification is too strong, the neural network is likely to reach 100\% channel sparsity at some point. When it happens, the training collapses as the cross-entropy loss becomes infinity. Therefore, we need to decease the values of penalty parameters earlier than the learning rate to stabilize the training.
\begin{figure}
    \centering
    \includegraphics[scale=0.7]{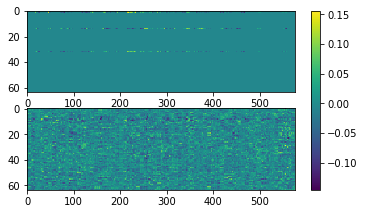}
    \caption{Visualization the 45th layer of a float resnet56 (bottom) and a 4-bit resnet56 pruned by APGDSSM (top). The layer originally has shape [64,64,3,3] and is permuted and reshaped to shape 64$\times$576 for visualization. Each row of the plots represents a channel.}
    \label{fig:layer}
\end{figure}
\section{Results}
We validate Algorithm 1 in CIFAR10 and CIFAR100 with ResNet (\cite{he2016deep}). The results are shown in Table \ref{tab:r1} and Table \ref{tab:r2}. As the tables show, the GL penalty and the shrinkage operator can significantly improve the weight sparsity and the channel sparsity with minor reduction on accuracy. The splitting step before the shrinkage operator can greatly improve the sparsity. Of course, the model performance would be somewhat affected. 
\begin{figure}
    \centering
    \includegraphics[scale=0.4]{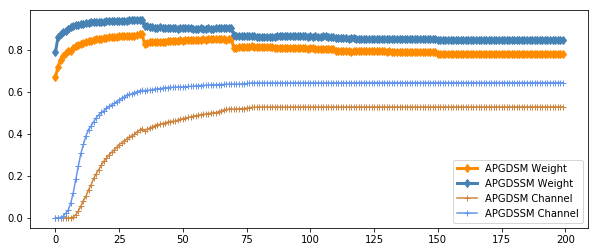}
    \caption{ Weight sparsity and channel sparsity against epochs. The weights sparsity deceases at milestones of the adaptive schemes, while the channel sparsity has smooth convergence}
    \label{fig:sparsity}
\end{figure}
\begin{table}[ht!]
    \centering
\begin{tabular}{c|c|c|c|c|c|c}
             Model  & Pruning &Ch. sp  & Wt .sp &Accuracy \\
    \hline 
    \hline
    Resnet20 & None&9.53\%&42.73\%&91.75\%\\
    \hline
     Resnet20 &APGDSM &14.67\%&72.68\%&91.53\%\\
     \hline
     Resnet20 &APGDSSM&24.56\%&85.04\%&90.64\%\\
     \hline
     \hline
     Resnet56 & None&25.16\%&61.83\%&93.24\%\\
    \hline
     Resnet56 & APGDSM &52.76\%&78.11\%&92.58\%\\
     \hline
     Resnet56 & APGDSSM&64.28\%&84.59\%&91.69\%\\
     \hline
     
\end{tabular}
\caption{4-bit quantized models with pruning methods on Cifar10 dataset. The initial values of parameters are $\lambda_1^1=0.04, \lambda_2^1=5.e-6, \beta^1 =1.e-3$.}
\label{tab:r1}
\end{table}
\begin{table}[ht!]
    \centering
\begin{tabular}{c|c|c|c|c}
             Model  & Pruning Method &Ch. sp  & Wt .sp &Accuracy \\
    \hline 
    \hline
    R.110 & None &24.63\%&53.20\%&71.74\%\\
    \hline
     R.110 & APGDSM &33.61\%&69.44\%&71.68\%\\
     \hline
     R.110 & APGDSSM &36.62\%&85.04\%&71.59\%\\
    \hline

\end{tabular}
\caption{4-bit quantized models with channel pruning methods on Cifar100. Initial values of parameters are $\lambda_1^1=0.02, \lambda_2^1=5e-6, \beta^1 = 1e-3 $. R.= Resnet.}
\label{tab:r2}
\end{table}

\begin{figure*}
    \centering
    \includegraphics[width=12cm, height=5cm]{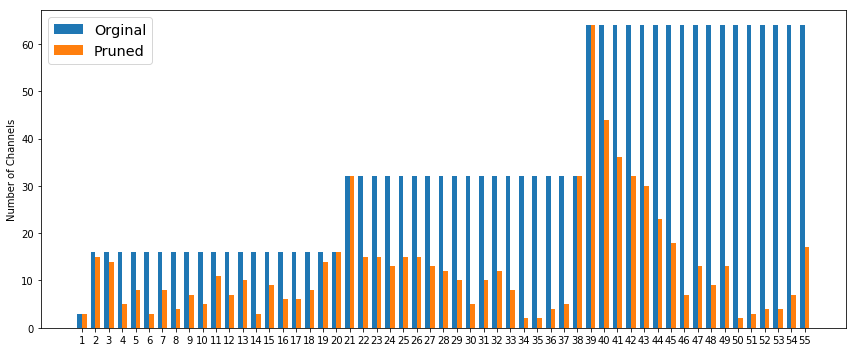}
    \caption{The blue bars are the numbers of channels in layers of float ResNet56. The orange bars are the numbers of channels in layers of pruned 4-bit model by APGDSSM.  The 55 of the 56 layers in ResNet56 are convolutional.}
    \label{fig:pruning}
\end{figure*}
Meanwhile, we numerically verify the convergence of the sparsity in Figure \ref{fig:sparsity}. Although the weight sparsity will decease every time the values of parameters updated, the channel sparsity has a nice convergence along training. The channel-wise GL penalty is the key to push the weight sparsity created by shrinkage and potential splitting into channel sparsity. In Figure \ref{fig:pruning}, we show the comparison  of a float ResNet56 and a 4-bit ResNet56 pruned via APGDSSM. We see that the channels are largely pruned in most layers. 

\section{efficiency and accuracy trade-off}
\begin{figure}[ht!]
    \centering
    \includegraphics[width=6cm, height=3.5cm]{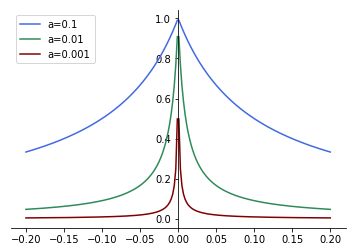}
    \caption{CT$l_1$ penalty $1-\frac{|x|}{a+|x|}$ for different values of $a$.}
    \label{fig:NTL}
\end{figure}

As we mentioned in the previous sections, the training might collapse if the adaptive scheme and parameter values are selected carelessly. In order to further compress the models, we propose a complementary transformed-$l_1$ (CTL$_1$) penalty to prevent the training from collapse. This penalty is inspired by the transformed $l_1$ (TL$_1$) regularization in robust compressed sensing \cite{zhang2014minimization}. We define
$$||x||_{CTL_1, a}:=1-\rho_a(x)=1-\frac{|x|}{a+|x|}$$
We remark that $||\cdot||_{CTL, a}$ is not a norm but only a regularization. We abuse the norm notation here for convenience. Note that 
$$\lim_{a\to 0^{+}}||x||_{CTL_1, a}=1-||x||_0=
\begin{cases}
1 & x = 0\\
0 & x \not = 0
\end{cases}
$$
For small choice of $a$, the value of $||x||_{CTL_1, a}$ is negligible when $|x|$ is large. The behavior of the CTL$_1$ penalty is illustrated in figure \ref{fig:NTL}. To prevent the neural network from having a zero layer, we apply it to each layer of our model
$$||\mathbf{w}||_{CTL_1,a}:=\sum_{l=1}^{L}1-\frac{||w_l||_1}{a+||w_l||_1}$$
By imposing this CTL$_1$ penalty, we force each layer to have some nonzero weights, so the training will not collapse. The augmented objective is 
\begin{equation}
    \min_{\mathbf{u}\in \mathcal{Q}}\mathcal{L}(\mathbf{u}):=f(\mathbf{u})+\lambda_2 ||\mathbf{u}||_{GL}+\lambda_3||\mathbf{u}||_{CTL_1}+\lambda_1||\mathbf{u}||_1
\end{equation}
and a Lasso regularization $\lambda_1||u||_1$ is implicitly imposed via the shrinkage operator as in the previous section. As a result, we can have more 'aggressive' choices for the values of parameters and the adaptive scheme to further pruning the neural networks. 
\begin{algorithm}
\caption{APDSSM with CT$l_1$ penalty}\label{alg:cap}
\hspace*{\algorithmicindent} \textbf{Input}: Float weights $w^0$. Hyperparameters $\lambda_1, \lambda_2, \beta$.  \\
\hspace*{\algorithmicindent} \textbf{Output} : Quantized weights $u$.
\begin{algorithmic}
\For{$t = 1,\cdots, 200$}:
    
    \State $\mathbf{u}^t = Proj_{Q}(\mathbf{w}^t)$
    \State $f(\mathbf{u}^t)=l(\mathbf{u}^t)+\gamma^t \lambda_2||\mathbf{u}^t||_{GL}+ \lambda_3 ||\mathbf{u}^t||_{CTL_1,\gamma^t a}$
    \State $\mathbf{w}^{t}=\mathbf{w}^{t-1}-\gamma^t\nabla f(\mathbf{u}^t)$ 
    \State $\mathbf{w}^t=\mathbf{w}^t-\gamma^t \beta (\mathbf{w}^t-\mathbf{u}^t)$
    
    \State $\mathbf{w}^{t}=Prox_{\gamma^t \lambda_1}(\mathbf{w}^{t-1}_g)$
\EndFor
\State $\mathbf{u} = Proj_{Q}(\mathbf{w}^{200})$
\end{algorithmic}
\end{algorithm}

\begin{table}[ht!]
    \centering
\begin{tabular}{c|c|c|c|c}
             Model  &$\lambda_2$ initial&Ch. sp  & Wt. sp &Accuracy \\
    \hline 
    \hline
    \multicolumn{4}{c}{Cifar10} \\
    \hline
    R.56  &$1.5\cdot 10^{-3}$&73.67\%&95.80\%&90.27\%\\
    \hline
     R.56  &$5\cdot 10^{-3}$&82.90\%&96.70\%&88.71\%\\
     \hline
     \multicolumn{4}{c}{Cifar100} \\
     \hline
     R.110 & $5\cdot 10^{-4}$&55.12\%&80.07\%&70.75\%\\
     \hline
     R.110 &$1\cdot10^{-3}$&58.06\%&80.75\%&70.16\%\\
    \hline

\end{tabular}
\caption{The stronger pruning scheme stabilized by CT$l_1$ penalty allows trade-off of a wider range of accuracy for efficiency, $\lambda_1=0.2$, $\beta = 0.01$; R.=Resnet.}
\label{tab:NTL}
\end{table}
\begin{table}[ht!]
    \centering
    \begin{tabular}{c|c|c|c|c}
          Pruning&Wt. sp& Ch. sp & Accuracy \\
    \hline
         None & 82.07\%&6.09\%&67.41\%\\
         \hline
         APGDSSM(w. CTL$_1$)&87.83\%&18.36\%&64.02\%
    \end{tabular}
    \caption{Pruning 4-bit Resnet18 on ImageNet (1K classes). We have $(\lambda_1,\lambda_2,\lambda_3, \beta)= (10^{-2},2 \cdot 10^{-4}, 1, 10^{-3}).$}
    \label{tab:imagenet}
\end{table}
In Algorithm2, we let the parameters $\lambda_1$, and $\lambda_2$, $\lambda_3$ and $\beta$ have the same adaptive scheme by multiply it by the learning rate. This scheme makes the parameters decrease slower. Hence, as shown in Table \ref{tab:NTL}, the channel sparsity increases significantly. The CT$l_1$ penalty allows us to further trader-off the performance to efficiency based on our needs. Finally, we present our results on ImageNet in Table \ref{tab:imagenet}. We increase the channel sparsity from 6.09\% to 18.36\%.
\section{Conclusion}
In this paper, we proposed APGDSSM to integrate the penalty based channel pruning and QAT. We remark that relaxations of QAT (\cite{Dockhorn, binaryrelax}) will lead to sub-optimal outcomes, because such methods search the sparse subspace first and then find local optimal quantized weights around the searched sparse weights. The two subspaces need to be searched concurrently from the beginning. We verifies that APGDSSM can deliver sparse quantized neural network with minor trader-off for performance. Further, we designed an auxiliary complementary transformed $l_1$ penalty to prevent training from collapsing, so we can trade more performance for efficiency if needed.

\newpage
\bibliographystyle{IEEEbib}
\bibliography{strings,refs}

\end{document}